\pdfoutput=1

\documentclass[11pt]{article}
   
\usepackage{ACL2023}

\usepackage{times}
\usepackage{latexsym}
\usepackage{amsmath}
\usepackage[most]{tcolorbox}
\usepackage{booktabs}    
\usepackage{multirow}    
\usepackage{graphicx}    
\usepackage{caption}     
\usepackage{booktabs}
\usepackage{multirow}
\usepackage[table]{xcolor}

\usepackage[T1]{fontenc}

\usepackage[utf8]{inputenc}

\usepackage{microtype}

\usepackage{inconsolata}
\usepackage{mathtools} 

\usepackage{amsmath}
\usepackage{amssymb}
\usepackage{graphicx}

\usepackage{listings}
\usepackage{xcolor}

\lstdefinestyle{jsonstyle}{
    backgroundcolor=\color{lightgray}, 
    basicstyle=\ttfamily\footnotesize,
    breaklines=true,
    frame=single,
    rulecolor=\color{black},
    showstringspaces=false,
    numbers=left,
    numberstyle=\tiny,
    stepnumber=1,
    numbersep=5pt,
    commentstyle=\color{gray},
    stringstyle=\color{red}
}

\usepackage[most]{tcolorbox}
\usepackage{xcolor}
\usepackage{amsmath}
\usepackage{enumitem}
\usepackage{needspace}

\definecolor{problemgray}{RGB}{246,246,246}
\definecolor{linegray}{RGB}{220,220,220}
\definecolor{pumfill}{RGB}{244,248,255}
\definecolor{prmfill}{RGB}{255,246,244}
\definecolor{takefill}{RGB}{248,248,248}

\tcbset{
  enhanced,
  boxrule=0.35pt,
  arc=1mm,
  left=1.8mm,
  right=1.8mm,
  top=1.2mm,
  bottom=1.2mm,
  fontupper=\small,
}

\newtcolorbox{problemstrip}{
  colback=problemgray,
  colframe=linegray,
  boxrule=0.35pt,
  arc=1mm
}

\newtcolorbox{pumstrip}{
  colback=pumfill,
  colframe=linegray,
  title=\textbf{PUM},
  fonttitle=\bfseries\small,
  coltitle=black
}

\newtcolorbox{prmstrip}{
  colback=prmfill,
  colframe=linegray,
  title=\textbf{PRM},
  fonttitle=\bfseries\small,
  coltitle=black
}

\newtcolorbox{takeawaystrip}{
  colback=takefill,
  colframe=linegray,
  boxrule=0.3pt,
  arc=1mm
}

\newcommand{\caseheading}[1]{%
  \vspace{1.2em}
  \noindent{\large\bfseries #1}\par\vspace{0.4em}
}

\title{From Correctness to Utility:\\ Gain-Based Prefix Evaluation for LLM Reasoning}

\author{
\textbf{Yuhang Zhou}\,$^{1,2}$ \quad
\textbf{Yixin Cao}\,$^{1,2\dagger}$ \quad
\textbf{Guangnan Ye}\,$^{1,2\dagger}$ \\[6pt]
$^{1}$Fudan University \quad
$^{2}$Shanghai Innovation Institute \\
\texttt{yuhangzhou22@m.fudan.edu.cn, yxcao@fudan.edu.cn, yegn@fudan.edu.cn}\\
}

\begin{document}
\maketitle


\renewcommand{\thefootnote}{\arabic{footnote}} 

\begin{abstract}

Reasoning prefixes shape the future trajectory of LLM problem solving, yet existing process reward models usually evaluate them through local step correctness. We argue that correctness is a useful but indirect proxy for the effect we ultimately care about: whether a prefix increases the probability of successful completion. We define this effect as prefix gain, the solve-rate improvement induced by conditioning lightweight student model group on a prefix, and use it to train a Prefix Utility Model (PUM) with a simple pairwise ranking objective. PUM learns outcome-grounded prefix utility and can score both complete trajectories and partial reasoning prefixes. Across Best-of-$N$ selection, beam search, and reinforcement learning on mathematical reasoning, PUM provides a strong prefix-level supervision signal, especially when candidate pools are large, search budgets increase, or rule-based rewards are sparse. We release all data, models, and code at \url{https://zhiqix.github.io/pum-project-page}.

\end{abstract}

\section{Introduction}

Reasoning with large language models (LLMs) is a sequential process in which intermediate prefixes shape future continuations~\cite{wei2022chain,wang2022self,yao2023tree}. A useful prefix can expose a promising decomposition and move the model toward a more solvable state, whereas a misleading prefix can bias the model toward an unproductive trajectory long before the final answer is produced~\cite{zhou2022least,liao2025lost,wang2025bounds}. This raises a central question for process-level supervision: \emph{what makes a reasoning prefix valuable?}

Prior work has approached prefix evaluation from two directions.
PRM-style methods provide dense step-level supervision by judging whether each intermediate step is locally valid or likely to follow a correct solution path~\cite{lightman2024let,wang2024math,zhang2025lessons}.
However, local correctness is not equivalent to prefix usefulness: a valid step may still leave the remaining trajectory brittle, while an incomplete or non-standard prefix may reveal a key idea that improves downstream solvability~\cite{liao2025lost,wang2025bounds}.
Recent rollout-based methods are more future-aware, estimating state values from continuations. Yet these signals are typically tied to a particular continuation policy or trajectory distribution, and do not explicitly measure the marginal improvement caused by the prefix itself~\cite{li2025process,setlur2025rewarding}.

\begin{figure}[t]
    \centering
    \includegraphics[width=0.45\textwidth]{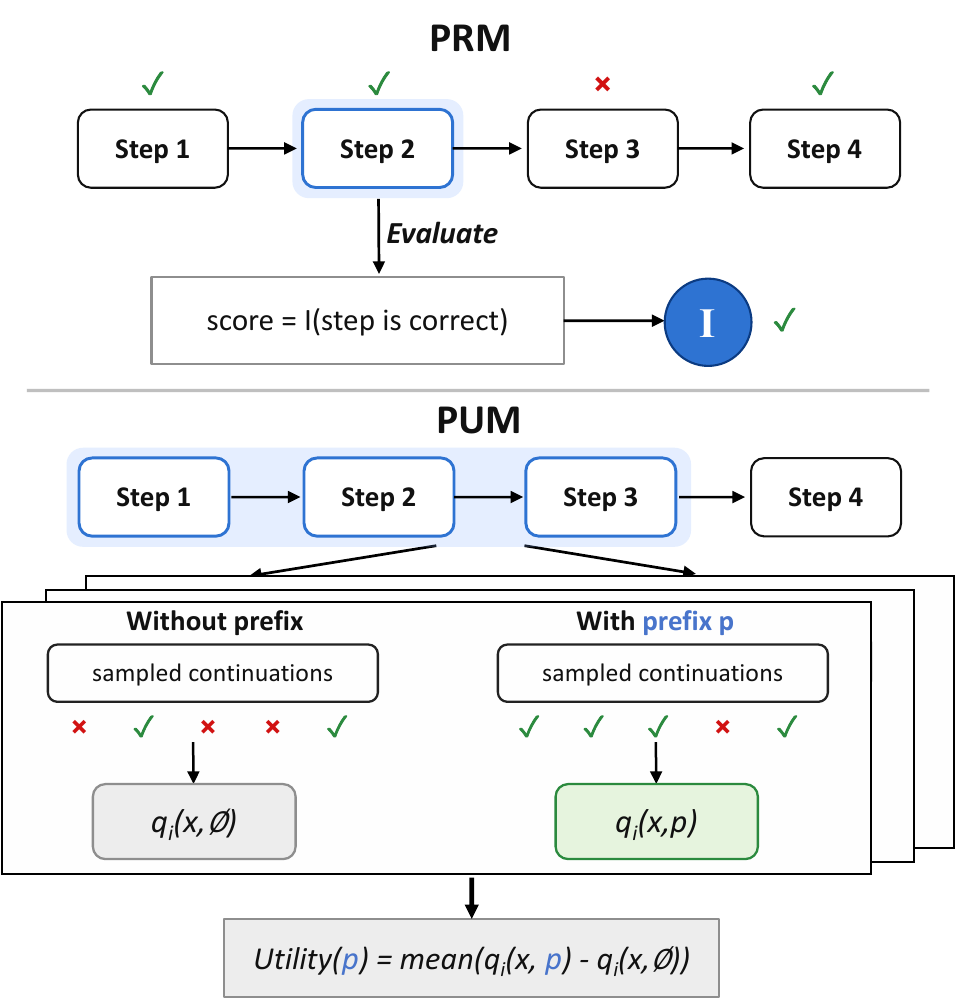}
    \caption{
    Correctness and Utility-based Evaluation.
    }
    \label{fig:intro}
\end{figure}

In this paper, we formulate prefix evaluation through \textbf{gain}: the marginal improvement in downstream solve probability induced by a prefix, as shown in Figure~\ref{fig:intro}.
Given a problem $x$ and a prefix $p$, we compare the probability of solving the problem after observing $p$ with that of solving it from scratch.
This with-prefix versus without-prefix comparison distinguishes gain from standard value-style evaluation: value estimates whether a prefix is likely to lead to success~\cite{li2025process}, whereas gain asks whether the prefix itself improves the chance of success relative to the same problem and solver population.
This marginal formulation better separates prefix usefulness from problem difficulty and from a solver's prior ability.

To estimate gain scalably, we use lightweight student models as probes. 
Each student solves the problem with and without the prefix, and the solve-rate difference yields a gain profile across capability levels. 
We decompose this profile into a shared intrinsic utility and policy-dependent variation, and use a simple scalarization: normalizing gains and averaging them over a reference population. 
Based on this framework, we construct PUM-Math from 20K mathematical reasoning trajectories and 280K outcome-grounded pairwise utility preferences, without manual step-level annotations. 
We then train a Prefix Utility Model (PUM), an LLM-based scalar evaluator with a value head, using a Bradley-Terry~\cite{bradley1952rank} loss.

We evaluate PUM on mathematical reasoning, where answer verification enables prefix-gain estimation. 
In controlled comparisons, we fix the backbone, MATH-origin data, and training recipe, and vary only the process-supervision target. 
Across Best-of-$N$ selection, beam search, and reinforcement learning (RL), PUM provides an effective prefix-utility signal, with the largest gains when candidate pools are large, search budgets increase, or rule-based rewards are sparse. 
Complete responses can be ranked as terminal prefixes, partial branches can be scored during search, and nested prefix-utility differences can provide dense RL advantages. 
Further analyses show that gain-based supervision supports weak-to-strong generalization, capability-aware aggregation, scaling, and more efficient construction than large-scale step-supervision datasets.

Our contributions are summarized as follows:
\begin{itemize}
\item We formulate prefix evaluation as counterfactual gain estimation, measuring how much a prefix improves downstream solve probability.
\item We construct PUM-Math, a gain-based prefix supervision dataset built from 20K reasoning trajectories and 280K outcome-grounded pairwise utility preferences. 
\item We train PUM on this data and show that gain-based prefix supervision improves Best-of-$N$ selection, beam search, and RL.
\end{itemize}

\section{Related Works}

Process-level supervision has become an important direction for improving multi-step reasoning. In mathematical reasoning and logical deduction, intermediate prefixes often shape the future trajectory: useful prefixes can guide later reasoning, while misleading ones may cause early lock-in~\cite{zhang2026bread,liao2025lost,wang2025bounds}. Process reward models (PRMs) therefore score intermediate steps to provide denser feedback than final-answer rewards~\cite{lightman2024let,zhang2025lessons}. However, most PRMs are trained as local correctness classifiers, which may not directly capture how a prefix changes future solvability~\cite{wang2024math}.

Recent work has explored more future-aware process evaluation. Process Q-value Models estimate intermediate-state values from an MDP perspective~\cite{li2025process}, while conditional reward models connect partial solutions to final outcomes~\cite{zhang2025linking}. Besides, PAV rewards step-wise progress using best-of-4 probing from the current policy model~\cite{setlur2025rewarding}. These methods move beyond local validity, but mainly focus on absolute state value, outcome-conditioned state quality, or policy-specific progress along a trajectory.

PUM instead evaluates arbitrary prefixes through outcome-grounded \emph{gain}. It measures the marginal contribution of a prefix by comparing downstream solve probability with and without the prefix, separating prefix usefulness from problem difficulty and solver prior ability. By estimating gain over a population of lightweight students, PUM represents prefix utility as a gain profile that captures both transferable intrinsic utility and policy-dependent variation, providing a unified prefix-level signal for selection, search, and credit assignment.

\begin{figure*}
    \centering
    \includegraphics[width=1\linewidth]{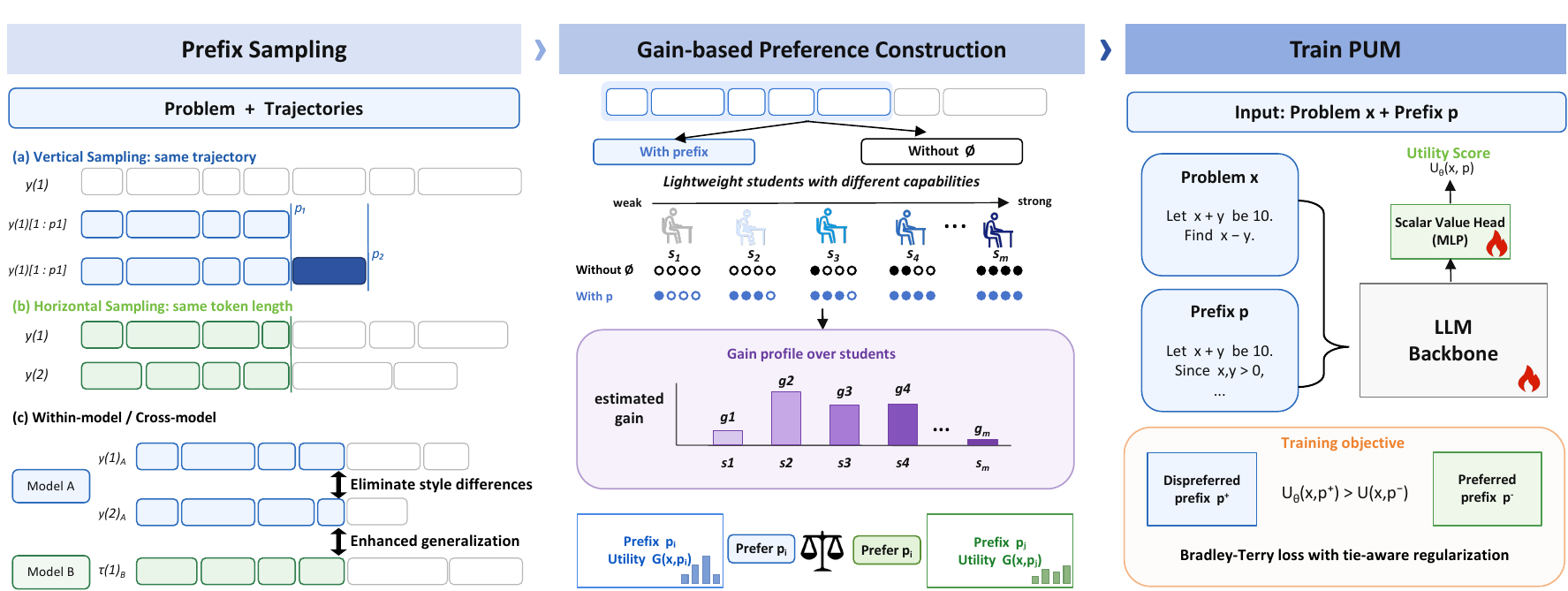}
    \caption{Prefix Utility Model: prefix sampling, gain-based preference construction, and pairwise utility learning.}
    \label{fig:main}
\end{figure*}

\section{Prefix Utility Model}
\label{sec:pum}

We introduce the Prefix Utility Model (PUM), an outcome-grounded evaluator for reasoning prefixes, as shown in Figure~\ref{fig:main}. We first sample diverse prefixes from trajectories, then estimate prefix-induced gains by comparing lightweight students' solve rates with and without each prefix, convert gain differences into pairwise preferences, and train PUM with a ranking objective.

\subsection{Prefix-Induced Gain Distribution}
\label{sec:prefix_gain}

Let \(x\) denote a reasoning problem, \(y\) a complete reasoning trajectory, and \(p\preceq y\) an arbitrary reasoning prefix before the final answer. Unlike step-level process supervision, \(p\) does not need to align with a manually segmented step boundary. Let \(V(x,y)\in\{0,1\}\) be an answer verifier that checks whether the final answer in \(y\) is correct.

For a student model \(s\) with policy \(\pi_s\), we define the prefix-conditioned success probability as
\begin{equation}
q_s(x,p)
=
\Pr_{c\sim \pi_s(\cdot\mid x,p)}
\left[
V(x,p\oplus c)=1
\right],
\end{equation}
where \(c\) is the continuation generated after conditioning on prefix \(p\). The corresponding no-prefix baseline is \(q_s(x,\emptyset)\), the probability that the same student solves the problem from scratch. We define the gain induced by prefix \(p\) for student \(s\) as
\begin{equation}
g_s(x,p)
=
q_s(x,p)-q_s(x,\emptyset).
\end{equation}
A positive gain means that the prefix increases the student's chance of reaching a correct final answer, while a negative gain means that it makes successful continuation less likely.

Inspired by Teach2Eval~\cite{zhouteach2eval}, we estimate \(g_s(x,p)\) by sampling continuations from a reference group of lightweight student models
\(\mathcal{S}=\{s_1,\ldots,s_M\}\), which cover different capability levels. For each prefix, these students produce a set of empirical gains:
\begin{equation}
\label{eq:gain_profile}
\hat{\mathbf{g}}(x,p)
=
\left[
\hat{g}_{s_1}(x,p),
\hat{g}_{s_2}(x,p),
\ldots,
\hat{g}_{s_M}(x,p)
\right].
\end{equation}

\subsection{From Gain Profile to Intrinsic Utility}
\label{sec:gain_structure}

The gain profile in Eq.~\eqref{eq:gain_profile} contains one gain estimate for each lightweight student. Before using this distributional profile as supervision, we first examine what structure it contains. If different students responded to prefixes in unrelated ways, then compressing their gains into a scalar target would not provide a meaningful learning signal.

We conduct a diagnostic study using lightweight student models with different initial solve rates, details are provided in Appendix ~\ref{app:empirical_study}. At the position-averaged level, gain curves are highly aligned across students. Prefixes from answer-correct trajectories tend to show increasing gain as reasoning progresses, while prefixes from answer-incorrect trajectories remain low or decrease. This suggests that student-measured gains contain a shared component that reflects the general usefulness of a prefix. At the individual-prefix level, however, agreement is more moderate, and students with closer baseline solve rates tend to assign more consistent gains. This indicates that the same prefix may affect different solvers differently, depending on their capability and reasoning behavior.

These observations motivate a two-component view of prefix utility. We define \emph{intrinsic utility} as the transferable contribution of a prefix that is shared across a group of solvers. It reflects whether the prefix provides useful information, a productive reasoning direction, or a helpful decomposition that generally makes the remaining problem easier. And define \emph{policy-dependent utility} as the solver-specific effect of the same prefix, which depends on a model's capability, prior knowledge, reasoning style, and sensitivity to format. For example, a prefix may provide large gain to a weaker student by revealing a missing strategy, while bringing little additional benefit to a stronger student that can already solve the problem unaided.

Based on this view, we interpret the empirical gain profile as a distribution over solver responses:
\begin{equation}
\hat{\mu}_{x,p}
=
\frac{1}{M}
\sum_{m=1}^{M}
\delta_{\hat{g}_{s_m}(x,p)} ,
\end{equation}
where \(\delta_a\) denotes a point mass at \(a\). The central tendency of this distribution estimates intrinsic utility, while its variation reflects policy-dependent utility.

In this work, we adopt the simplest scalar compression for training PUM. Since different students have different baseline abilities and gain scales, we normalize gains separately for each student and define
\begin{equation}
G_{\mathrm{intr}}(x,p)
=
\frac{1}{M}
\sum_{m=1}^{M}
\mathrm{Norm}_{m}(\hat{g}_{s_m}(x,p)).
\end{equation}
This scalar target estimates the shared, transferable component of prefix utility across the student group. It does not exhaust the full gain profile; rather, it provides a stable supervision target for the main PUM, while capability-aware variants can be obtained by assigning non-uniform weights to different students.

\subsection{Gain-Based Pairwise Preference Construction}
\label{sec:pairwise_construction}

Directly regressing scalar utility is undesirable because gain values are not directly comparable across problems and small differences can be dominated by rollout noise. We therefore convert intrinsic utility targets into within-problem pairwise preferences, so that supervision focuses on relative prefix usefulness under the same problem context.

For two prefixes \(p_i\) and \(p_j\) from the same problem \(x\), we compare their intrinsic utility estimates and assign a preference label according to an adaptive margin. The margin accounts for the reliability of the utility difference, and pairs with inconsistent student-wise preference directions are filtered out. As a result, a pair is labeled as \(p_i \succ p_j\), \(p_j \succ p_i\), tied, or discarded when the evidence is insufficient. Details of the adaptive thresholding and conflict-aware filtering procedure are provided in Appendix~\ref{app:pairwise_preference_construction}.

We construct candidate pairs from both vertical and horizontal comparisons. Vertical pairs compare prefixes at different truncation positions along the same trajectory, encouraging PUM to learn how utility evolves with reasoning progress. Horizontal pairs compare length-matched prefixes from different trajectories of the same problem, reducing reliance on prefix length and generation style.

This construction yields a gain-based preference dataset
\begin{equation}
\mathcal{D}_{\mathrm{pair}}
=
\{(x,p_i,p_j,y_{ij})\},
\end{equation}
where \(y_{ij}\) denotes a preference or tie label. The labels are derived from outcome-grounded solve-rate gains rather than manual step-level correctness annotations.

\subsection{Learning PUM by Pairwise Ranking}
\label{sec:pum_training}

PUM is a scalar utility model \(U_\theta(x,p)\) that predicts the intrinsic utility of a problem-prefix pair. In our implementation, PUM adds an MLP value head on top of a language model backbone. The input is formatted as the concatenation of the problem and the reasoning prefix, and the hidden state of the last non-padding token is used to produce a scalar utility score.

We train PUM with a pairwise Bradley-Terry objective~\cite{bradley1952rank}. Each training example contains two prefixes \((p_a,p_b)\) from the same problem and a label \(y\in\{1,0,-1\}\), where \(1\) means \(p_a\) is preferred, \(-1\) means \(p_b\) is preferred, and \(0\) denotes a tie. We compute
\begin{equation}
\Delta_\theta
=
U_\theta(x,p_a)-U_\theta(x,p_b).
\end{equation}
The label is mapped to a soft target
\begin{equation}
t =
\begin{cases}
1.0, & y=1,\\
0.5, & y=0,\\
0.0, & y=-1,
\end{cases}
\end{equation}
and the training loss is
\begin{equation}
\mathcal{L}_{\mathrm{PUM}}
=
-\mathbb{E}
\left[
t\log\sigma(\Delta_\theta)
+
(1-t)\log(1-\sigma(\Delta_\theta))
\right].
\end{equation}
Thus, tie pairs are trained as soft preferences with target \(0.5\), encouraging the model not to assign a strong ordering when the gain evidence is insufficient.

\section{Experiments}

\begin{figure*}[t]
    \centering
    \includegraphics[width=0.98\textwidth]{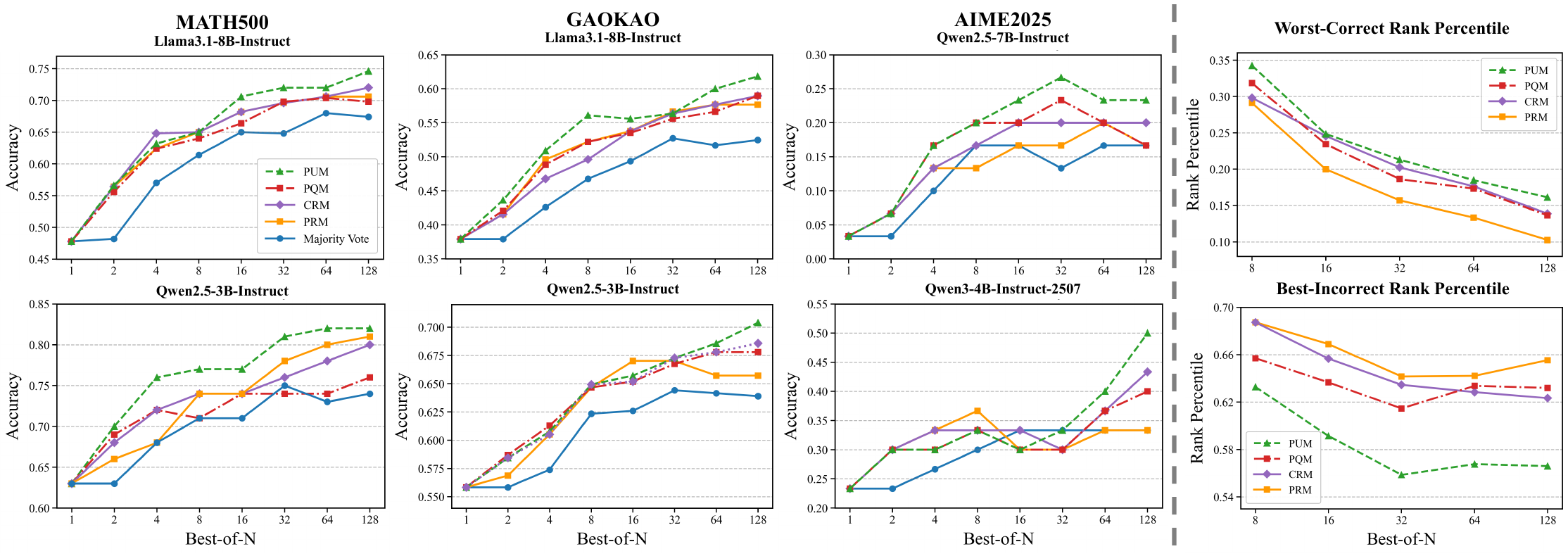}
    \caption{
    Best-of-N selection and ranking robustness.
    Left: Best-of-N selection accuracy.
    Right: ranking robustness, where higher worst-correct and lower best-incorrect percentiles indicate better evaluator behavior.
    }
    \label{fig:bon}
\end{figure*}

\subsection{Experimental Setup}
\label{sec:exp_setup}

\paragraph{PUM training.}
Following prior process-supervision work, we instantiate PUM on mathematical reasoning, where reliable final-answer verification enables controlled estimation of prefix-induced solve-rate gains. To enable a fair comparison with MATH-Shepherd~\cite{wang2024math} style process supervision, we construct our training data from the MATH training set. Following the procedure in Section~\ref{sec:pairwise_construction}, we obtain \(280\mathrm{k}\) gain-based pairwise data, detailed in Appendix~\ref{app:estimating_prefix_utility}. For PUM, we use Qwen3-4B-Instruct-2507~\cite{yang2025qwen3} as the backbone, details in Appendix~\ref{app:pum_training_details}. 

\paragraph{Baselines and evaluation protocols.}
We compare PUM with several process-level evaluators, including vanilla PRM~\citep{lightman2024let}, PQM~\citep{li2025process} and CRM~\cite{zhang2025linking}. 
Different process evaluators require different forms of supervision. PRM and CRM rely on step-level correctness annotations, and we therefore use Math-Shepherd~\cite{wang2024math} supervision for these baselines. PQM and PUM are trained on the same prefix trajectory pool, but differ in their supervision targets: PQM learns value-style process scores, while PUM learns gain-based prefix utility. This setup allows us to compare established evaluator pipelines under their standard supervision regimes, while using the PQM-PUM comparison to isolate the effect of value-based versus gain-based prefix supervision.

\subsection{Best-of-N Selection}
\label{sec:bon}

Best-of-N selection evaluates whether an evaluator can select the most promising trajectory from \(N\) sampled responses. For PUM, each complete response is treated as a terminal prefix and ranked by its predicted utility. We evaluate four policy models, Qwen2.5-3B-Instruct, Qwen2.5-7B-Instruct~\cite{bai2023qwen}, Qwen3-4B-Instruct-2507~\cite{yang2025qwen3} , and Llama-3.1-8B-Instruct~\cite{touvron2023llama}, on GAOKAO2023~\cite{liao2024mario}, MATH500~\cite{hendrycks2021measuring}, and AIME2025 benchmarks with \(N\in\{1,2,4,8,16,32,64,128\}\). Detailed scoring definitions are provided in Appendix~\ref{app:best_of_n_details}.

As shown in Figure~\ref{fig:bon} left, PUM achieves the best or near-best Best-of-N accuracy across policy models and datasets. Its advantage is most visible at larger candidate pools, where the evaluator must identify genuinely correct trajectories among many plausible but incorrect responses. Majority voting often saturates when correct answers do not dominate the sample distribution, and PRM-style baselines may over-score locally plausible reasoning. By ranking complete responses with outcome-grounded prefix utility, PUM provides a stronger trajectory-level verification signal, especially on MATH500 and GAOKAO2023. Although AIME results are noisier due to higher difficulty and fewer problems, PUM remains competitive and often performs best in large-\(N\) settings.

\begin{table*}[t]
\centering
\small
\setlength{\tabcolsep}{6.0pt}
\renewcommand{\arraystretch}{1.12}
\caption{Beam Search accuracy on MATH500 and GAOKAO2023.}
\label{tab:beam_search}
\begin{tabular}{llcccccccc}
\toprule
\multirow{2}{*}{Models} 
& \multirow{2}{*}{Methods}
& \multicolumn{4}{c}{MATH500}
& \multicolumn{4}{c}{GAOKAO2023} \\
\cmidrule(lr){3-6} \cmidrule(lr){7-10}
& 
& \(N=4\) & \(N=8\) & \(N=20\) & \(N=100\)
& \(N=4\) & \(N=8\) & \(N=20\) & \(N=100\) \\
\midrule

\multirow{4}{*}{\begin{tabular}[c]{@{}l@{}}Qwen2.5-3B\\-Instruct\end{tabular}}
& PRM & 66.40 & \underline{68.40} & 72.80 & 69.20
      & \underline{58.70} & 60.26 & 61.56 & 42.34 \\
& PQM & 65.60 & 66.00 & 68.20 & 66.00
      & 56.88 & 61.30 & 63.89 & 65.45 \\
& CRM & \underline{67.60} & 68.20 & \underline{73.80} & \underline{74.60}
      & 57.92 & \underline{63.38} & \underline{64.94} & \underline{66.88} \\
& PUM & \textbf{71.80} & \textbf{71.80} & \textbf{77.60} & \textbf{78.00}
      & \textbf{61.04} & \textbf{67.01} & \textbf{68.57} & \textbf{71.43} \\
\midrule

\multirow{4}{*}{\begin{tabular}[c]{@{}l@{}}Llama3.1-8B\\-Instruct\end{tabular}}
& PRM & 43.20 & 44.00 & 40.00 & 23.00
      & 38.96 & 37.92 & 41.04 & 27.01 \\
& PQM & 42.80 & 45.00 & 48.20 & 53.00
      & 35.06 & 39.74 & 45.19 & 43.90 \\
& CRM & \underline{50.20} & \underline{53.20} & \underline{57.80} & \underline{65.60}
      & \underline{41.30} & \underline{42.86} & \underline{47.79} & \underline{53.25} \\
& PUM & \textbf{57.40} & \textbf{60.00} & \textbf{68.00} & \textbf{73.60}
      & \textbf{44.94} & \textbf{50.91} & \textbf{55.06} & \textbf{63.90} \\

\bottomrule
\end{tabular}
\end{table*}

Beyond top-1 selection accuracy, we further evaluate evaluator robustness through rank percentiles. A robust evaluator should not only select the correct response at top-1, but also assign consistently high ranks to correct candidates and low ranks to incorrect ones. We therefore measure the worst correct rank percentile and the best incorrect rank percentile. The former should be higher, indicating that even the least-favored correct response remains relatively well ranked; the latter should be lower, indicating that the most-favored incorrect response is not ranked too highly. As shown in Figure~\ref{fig:bon} right, PUM achieves the best behavior on both metrics, maintaining higher ranks for correct responses while suppressing high-scoring incorrect responses. This indicates that PUM provides a more robust trajectory-level ranking signal than correctness-oriented baselines.

\subsection{Beam Search}

Beam Search experiments use the policy model to iteratively generate multiple candidate continuations from reasoning prefixes, while the evaluator scores and prunes beams at each search step. This setting evaluates the evaluator's ability to guide reasoning at the prefix level by identifying promising partial trajectories before the final answer is produced. For each question, beam search initiates by sampling N responses. Subsequently, a beam of b candidates with the highest rewards is maintained and expanded during the generation process.

Table~\ref{tab:beam_search} reports beam-search accuracy on MATH500 and GAOKAO2023. PUM consistently achieves the best performance across both datasets, both policy models, and all search budgets. The advantage becomes more pronounced as the search budget increases, suggesting that PUM is better at preserving partial reasoning prefixes that remain useful after further expansion. Notably, several baselines degrade as \(N\) increases, indicating that larger search budgets can amplify evaluator misalignment: high-scoring prefixes are not necessarily those that lead to correct final answers. On MATH500, PUM reaches \(78.00\%\) with Qwen and \(73.60\%\) with Llama at \(N=100\). On GAOKAO2023, PUM similarly achieves the best accuracy, reaching \(71.43\%\) and \(63.90\%\) under the two policy models. These results show that gain-based prefix utility provides an effective search-time signal for pruning incomplete reasoning branches before final answers are produced.

\subsection{Reinforcement Learning}

RL provides a stricter test of process evaluators than inference-time selection, because inaccurate rewards can be directly exploited by the policy. We combine evaluator feedback with rule-based GRPO~\cite{shao2024deepseekmath} through a two-level credit assignment design. The rule-based reward provides a trajectory-level signal based on final-answer correctness, anchoring optimization to the standard RLVR objective. The evaluator provides a process-level signal inside each rollout. For PUM, this signal is computed from nested prefix-utility differences, measuring whether a newly generated reasoning segment increases the utility of the current reasoning state. For PRM~\cite{lightman2024let} and PQM~\cite{li2025process}, we construct analogous dense process-level advantages from their step or value scores, with details in Appendix~\ref{app:rl_details}. CRM~\cite{zhang2025linking} is not included in this comparison because it does not naturally provide step-wise scores that can be converted into per-segment advantages under this design. All RL experiments are conducted with Qwen2.5-Math-7B as the policy model.

\begin{table*}[t]
\centering
\small
\setlength{\tabcolsep}{6.0pt}
\renewcommand{\arraystretch}{1.12}
\caption{
Pass@1 accuracy on mathematical reasoning benchmarks after RL training.
}
\label{tab:math_rl_results}
\begin{tabular}{llccccc}
\toprule
Setting & Method & MATH500 & AMC23 & Minerva Math & AIME(24-26) & AVG \\
\midrule

\multirow{5}{*}{Outcome-Anchored}
& Vanilla GRPO      & 76.60 & 52.50 & \textbf{29.04} & \underline{16.67} & 55.21 \\
& PRM+GRPO   & 75.40 & 60.00 & 25.37 & 14.44 & 53.55 \\
& PQM+GRPO         & 71.40 & 60.00 & 19.85 & 10.00 & 49.22 \\
& PUM+GRPO         & \textbf{80.40} & \textbf{70.00} & \underline{28.68} & \textbf{17.78} & \textbf{58.09} \\

\midrule

\multirow{3}{*}{Process-Only}
& PRM              & 69.80 & \underline{57.50} & 18.75 & 10.00 & 47.89 \\
& PQM              & \underline{70.00} & \underline{57.50} & \underline{20.22} & \underline{13.33} & \underline{48.78} \\
& PUM              & \textbf{80.20} & \textbf{65.00} & \textbf{26.10} & \textbf{20.00} & \textbf{57.21} \\

\bottomrule
\end{tabular}
\end{table*}

\begin{figure}[t]
    \centering
    \includegraphics[width=0.49\textwidth]{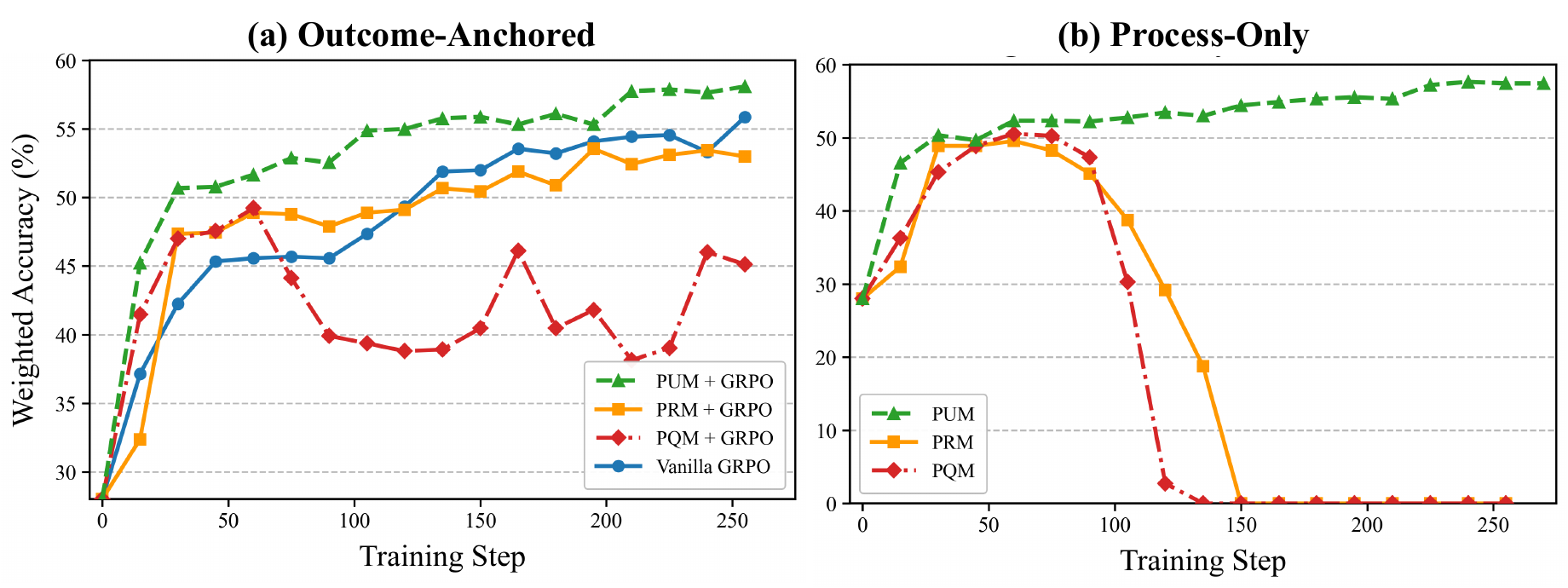}
    \caption{
    RL training curves.
    }
    \label{fig:RL}
\end{figure}

Figure~\ref{fig:RL} and Table~\ref{tab:math_rl_results} summarize the RL results. In the outcome-anchored setting, where rule-based GRPO is used as the final-answer reward, PUM+GRPO achieves the best overall performance, improving the weighted average accuracy from \(55.21\%\) for vanilla GRPO to \(58.09\%\). Compared with PRM- and PQM-based dense rewards, PUM yields a more stable improvement curve, suggesting that prefix-utility differences provide more reliable process-level credit assignment.

The process-only setting further exposes the robustness of different evaluators. As shown in Figure~\ref{fig:RL}, PRM and PQM training initially improves but later collapses, while PUM remains stable and reaches \(57.21\%\) average accuracy. We find that PRM and PQM exhibit stronger length preferences: the policy learns to generate longer responses that receive high process scores, while validation accuracy drops. This indicates reward hacking, where local process rewards are exploited without improving final-answer correctness.

The training dynamics also explain why PUM+GRPO is effective. In early training, rule-based GRPO provides sparse group-level signals because most rollout groups contain few correct answers. At this stage, the average absolute magnitude of PUM advantages is about \(10\times\) that of GRPO advantages, supplying dense guidance for cold start. As the policy improves and correct answers become more frequent, the GRPO signal strengthens, while PUM continues to provide fine-grained prefix-level credit assignment. This complementary behavior allows PUM and GRPO to jointly improve both early learning efficiency and final performance.

\begin{figure*}[t]
    \centering
    \includegraphics[width=0.98\textwidth]{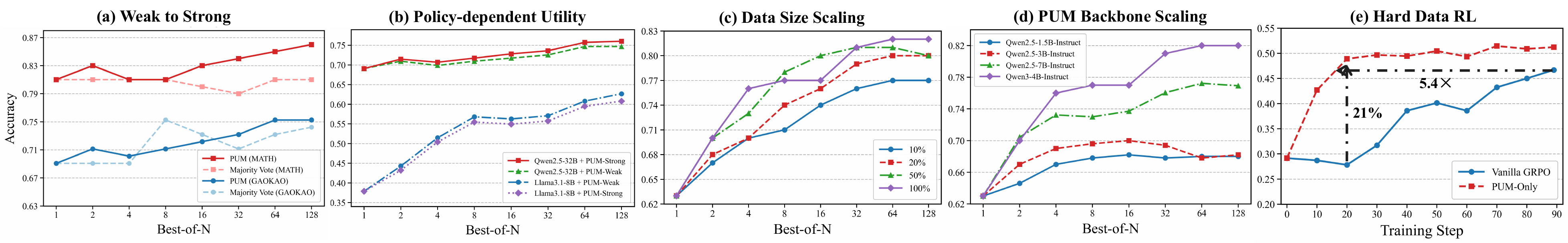}
    \caption{
   Further analyses on weak-to-strong, policy dependence, scaling, and hard-data RL.
    }
    \label{fig:further}
\end{figure*}

\subsection{Further Analysis}
\label{sec:further_analysis}

We conduct additional analyses to better understand the properties of evaluators trained from prefix utility.

\paragraph{Weak-to-strong generalization.}
We further evaluate whether PUM can generalize to a stronger policy model than those used for data construction and PUM backbone. We use Qwen2.5-32B-Instruct as the policy model, which is larger and stronger than the trajectory generators, student models, and PUM backbone. As shown in Figure~\ref{fig:further} (a), PUM still improves with larger candidate pools on both GAOKAO and MATH, reaching its best performance at large \(N\). This suggests that PUM captures an intrinsic prefix utility that transfers beyond the weaker models used to construct its supervision, rather than merely fitting their generation patterns.

\paragraph{Policy-dependent utility.}
We also study whether student weighting can adapt PUM to downstream policies of different strengths. We group student models by their initial solve rates and train PUM-Weak with weights \([0.6,0.3,0.1]\) and PUM-Strong with weights \([0.1,0.3,0.6]\) over weak, medium, and strong students. As shown in Figure~\ref{fig:further} (b), PUM-Weak performs better on the weaker Llama3.1-8B-Instruct policy, while PUM-Strong performs better on the stronger Qwen2.5-32B-Instruct policy. This suggests that, beyond intrinsic utility, prefix gain also contains a policy-dependent component that can be exploited by capability-aware aggregation.

\paragraph{Data scaling.}
We further study how PUM benefits from increasing amounts of gain-based supervision. We train PUM with different fractions of the pairwise utility data, including \(10\%\), \(20\%\), \(50\%\), and \(100\%\), and evaluate Best-of-N selection performance. As shown in Figure~\ref{fig:further} (c), larger training data consistently improves PUM, especially under larger candidate pools where the evaluator must distinguish fine-grained utility differences among responses. The \(100\%\) data setting achieves the best overall performance, while the \(50\%\) setting already approaches it at large \(N\), suggesting that gain-based supervision is data-efficient but still benefits from additional preference pairs, as even half of the data already recovers most of the full-data performance.

\begin{table*}[t]
\centering
\small
\setlength{\tabcolsep}{4.8pt}
\renewcommand{\arraystretch}{1.15}
\caption{
Supervision construction cost comparison.
PUM-Math avoids human step annotation and requires substantially less automatic labeling computation than Math-Shepherd.
}
\label{tab:pum_cost_comparison}
\resizebox{\textwidth}{!}{
\begin{tabular}{lcccccc}
\toprule
\textbf{Dataset} 
& \textbf{Domain}
& \textbf{\#Traj.}
& \textbf{Supervision}
& \textbf{Annotator / Labeler}
& \textbf{\#Calls}
& \textbf{Rel. Compute} \\
\midrule

PRM800K 
& MATH
& $\sim$75K
& 800K step labels
& Human annotators
& -- 
& Human labeling \\

Math-Shepherd
& MATH + GSM8K
& $\sim$445K
& $\sim$4.0M outcome-derived step labels
& 7B completer
& $\sim$35.6M
& $\mathbf{18.5\times}$ \\

\textbf{PUM-Math}
& MATH
& \textbf{$\sim$20K}
& \textbf{280K pairwise utilities}
& \textbf{Avg. 2B students}
& \textbf{6.72M}
& \textbf{1.0$\times$} \\

\bottomrule
\end{tabular}
}
\end{table*}

\paragraph{Backbone scaling.}

We analyze whether PUM performance is limited by the gain-based data or by the backbone model capacity. We train PUM with several Qwen-family backbones, including Qwen2.5-1.5B, Qwen2.5-3B, Qwen2.5-7B and Qwen3-4B,  using the same pairwise utility data. As shown in Figure~\ref{fig:further} (d), performance improves as the backbone becomes stronger. This scaling trend suggests that our data are not saturated under the current training setup, and that stronger base models can better learn the subtle utility distinctions encoded in prefix-gain supervision.

\paragraph{Learning from hard data.}
We further test PUM on hard problems where RLVR signals are extremely sparse. We collect \(5.7\mathrm{k}\) problems from DAPO-Math-17K for which the base model fails to produce any correct solution in \(16\) rollouts. In this setting, native GRPO receives almost no effective group-level advantage, while PUM can still provide dense supervision through utility advantage.

As shown in Figure~\ref{fig:further} (e), PUM-only training yields much faster improvement than vanilla GRPO. At step \(20\), PUM-only achieves around \(21\%\) absolute improvement over GRPO and reaches a performance level that GRPO approaches only after about \(5.4\times\) more training steps. This demonstrates that prefix-utility supervision substantially improves data efficiency on hard reasoning problems where rule-based rewards are sparse.

\paragraph{Supervision cost.}
We further compare the supervision construction cost of PUM-Math with representative process-supervision datasets. As shown in Table~\ref{tab:pum_cost_comparison}, PUM-Math avoids manual step-level annotation and is built from about 20K trajectories, much fewer than the 75K trajectories in PRM800K and the 445K trajectories in Math-Shepherd. Although PUM-Math evaluates multiple prefixes from each trajectory, its construction requires about 6.72M student rollouts, obtained from seven prefixes per trajectory and 48 continuations per prefix, using lightweight student models with an average size of 2B parameters. By contrast, Math-Shepherd constructs step-level supervision at a much larger scale: under a simple estimate with 10 reasoning steps per trajectory and \(N=8\) completions per step, it requires about 35.6M completions generated by a 7B model. 
We measured the average continuation length for both PUM-Math and Math-Shepherd and estimated labeling cost by parameter-token computation. Under this accounting, Math-Shepherd requires roughly \(18.5\times\) more labeling computation than PUM-Math, suggesting that PUM's gains come from more effective supervision targets rather than greater supervision cost.

\section{Conclusion}
\label{sec:conclusion}

We presented PUM, a gain-based prefix evaluator that shifts process evaluation from local correctness to outcome-grounded utility. By estimating solve-rate gains with lightweight student models, PUM learns an intrinsic prefix-utility target from gain-based pairwise preferences. Experiments show that this signal improves Best-of-$N$ selection, beam search, and reinforcement learning by supporting stronger verification, search-time guidance, and dense credit assignment. Further analyses demonstrate transfer to stronger policies, capability-dependent variation, data and backbone scaling, improved hard-data learning, and lower construction cost than large-scale step-supervision datasets. These results suggest that outcome-grounded gain provides an effective and efficient supervision target for prefix evaluation.

\section{Limitations}
\label{sec:limitations}

Our experiments focus on mathematical reasoning, where reliable final-answer verification enables controlled estimation of prefix gain and provides a clean testbed for comparing process-supervision targets. Extending PUM to domains with ambiguous, delayed, or human-judged outcomes would require suitable outcome signals or learned verifiers.

While PUM reduces construction cost compared with prior automated supervision pipelines, its gain estimation still requires substantial computation from student rollouts. In addition, this work adopts a simple scalarization of gain distributions by averaging normalized student gains into an intrinsic utility target, which enables a clean comparison with PRM-style supervision but does not fully exploit the distributional structure of prefix utility. Future work could explore richer distributional modeling, policy-adaptive utility estimation, and extensions to agentic settings, where intermediate plans, tool-use traces, or workspace states may be evaluated by their contribution to eventual task success.

\bibliography{anthology,custom}
\bibliographystyle{acl_natbib}

\clearpage
\appendix

\section{Prefix Utility Data Construction}

\begin{table*}[t]
\centering
\caption{
Trajectory generation models used for constructing PUM supervision. We use seven models from five organizations, yielding 20K trajectories in total. Thinking traces are removed for thinking-oriented models, and only the final answer-facing solution is retained as the trajectory.
}
\small
\setlength{\tabcolsep}{5.2pt}
\renewcommand{\arraystretch}{1.08}
\begin{tabular}{lccccc}
\toprule
Model & Size & Thinking & Organization & Max Tokens & \#Traj. \\
\midrule
MiniCPM4.1-8B & 8B & No & OpenBMB & 4096 & 3K \\
MiniCPM3-4B & 4B & No & OpenBMB & 4096 & 3K \\
Mimo-7B-RL & 7B & Yes & Xiaomi & 8192 & 2K \\
Hunyuan-1.8B-Instruct & 1.8B & No & Tencent & 4096 & 2K \\
Phi-4 & 14B & No & Microsoft & 4096 & 4K \\
Phi-3.5-mini-instruct & 4B & No & Microsoft & 4096 & 4K \\
Gemma-4-E4B & 8B & Yes & Google & 8192 & 2K \\
\midrule
Total / Avg. & 6.7B & -- & 5 orgs. & -- & 20K \\
\bottomrule
\end{tabular}
\label{tab:trajectory_generators}
\end{table*}

\subsection{Trajectory Generation}
\label{app:trajectory_generation}

To construct a diverse pool of reasoning trajectories, we use seven open models from five organizations as trajectory generators. The models cover different parameter scales, model families, and generation styles, with an average size of approximately 6.7B parameters. This diversity is important for reducing the risk that PUM learns model-specific stylistic artifacts rather than prefix utility. In particular, trajectories generated by models from different organizations expose PUM to heterogeneous formats, answer styles, and error patterns, while still keeping the overall data construction cost moderate.

Table~\ref{tab:trajectory_generators} summarizes the trajectory generation models. For all models, we use stochastic decoding with temperature $1.0$ and top-$p=1.0$. The maximum generation length is set to 4096 tokens for standard instruction models and 8192 tokens for thinking models. For thinking models, we remove the explicit thinking trace and retain only the final answer-facing solution as the trajectory. This normalization keeps the trajectory format comparable across generators and prevents PUM from relying on model-specific hidden-thinking markers.

The resulting trajectory pool contains 20K solutions in total. We deliberately include both smaller and stronger generators: weaker models contribute diverse failure cases and partially useful prefixes, while stronger models provide higher-quality reasoning trajectories. This mixture enables the subsequent prefix sampling procedure to compare prefixes across different quality levels, lengths, and reasoning styles, which is essential for constructing robust gain-based pairwise preferences.

\subsection{Prefix Sampling Strategy}
\label{app:prefix_sampling_strategy}

We construct prefix pairs in two stages: trajectory-pair sampling and prefix-pair sampling. The overall design is summarized in Table~\ref{tab:prefix_pair_types}.

\begin{table*}[t]
\centering
\caption{
Prefix-pair construction strategy. Vertical and horizontal pairs control prefix-level comparison, while within-model and cross-model pairs control trajectory-source variation.
}
\small
\setlength{\tabcolsep}{5.0pt}
\renewcommand{\arraystretch}{1.08}
\begin{tabular}{lll}
\toprule
Pair Type & Definition & Purpose \\
\midrule
Vertical 
& Same trajectory, different truncation positions 
& Learn utility progression \\
Horizontal 
& Same token length, different trajectories 
& Mitigate length shortcut and  bias \\
Within-model 
& Prefixes from the same generator 
& Reduce style confound \\
Cross-model 
& Prefixes from different generators 
& Improve style generalization \\
\bottomrule
\end{tabular}
\label{tab:prefix_pair_types}
\end{table*}

\paragraph{Trajectory-pair sampling.}

Following prior process-supervision work, we use problems from the MATH\cite{hendrycks2021measuring} training set as the base problem pool for trajectory construction. For each problem, we collect multiple reasoning trajectories generated by different models and form trajectory pairs under two settings: within-model and cross-model. Within-model pairs are sampled from the same generator, and we require the two trajectories to have different final-answer correctness, i.e., one correct and one incorrect. Cross-model pairs are sampled from different generators for the same problem, introducing variation in reasoning style and model-specific error patterns. We keep within-model and cross-model pairs balanced with a $1{:}1$ ratio.

\paragraph{Prefix sampling.}
Given the sampled trajectory pairs, we construct two types of prefix-level comparisons. For vertical pairs, we truncate the same trajectory at adjacent relative positions:
\[
\mathcal{R}=\{0.1,0.2,0.35,0.5,0.7,0.9\}.
\]
The $0.9$ position is taken before the final answer, i.e., it keeps most of the reasoning process while removing the answer itself. This prevents PUM from directly inferring correctness from the final answer.

For horizontal pairs, we compare prefixes from different trajectories. We first truncate each trajectory at the same set of relative positions, then retain a pair only when the two prefixes have similar token lengths:
\[
\frac{\left||p_i|-|p_j|\right|}{\max(|p_i|,|p_j|)} \leq 0.05.
\]
This length-matching constraint ensures that horizontal comparisons focus on reasoning quality rather than prefix length. 

In practice, we observe that PUM trained with these discrete truncation ratios still generalizes to prefixes truncated at unseen positions, indicating that the model does not simply memorize the predefined sampling grid.

All retained prefixes are later evaluated through student-model rollouts to estimate prefix-induced gain. Therefore, trajectory correctness is only used to sample informative trajectory pairs; the final preference labels are determined by gain-based utility differences rather than by trajectory correctness directly.

\subsection{Estimating Prefix Utility}
\label{app:estimating_prefix_utility}

We estimate prefix utility by measuring how much a prefix improves the downstream solve rate of lightweight student models. We use six students for gain estimation: Qwen3-1.7B, InternLM2.5-1.8B-Chat, Phi-4-mini, Gemma-E2B, Qwen3-0.6B, and SmolLM3-3B. For each problem-prefix pair, each student samples $K=8$ continuations. The no-prefix baseline for the same problem is estimated with the same rollout budget.

For continuation generation, we use temperature $0.7$ and top-$p=0.95$. We set the total context budget to 8192 tokens. Given a prefix $p$, the maximum continuation length is set to $8192-|p|$, so that longer prefixes receive shorter continuation budgets while all prefix-conditioned rollouts share the same total token budget.

For a student model $s$, the prefix-conditioned solve rate is estimated as
\[
\hat q_s(x,p)=\frac{1}{K}\sum_{k=1}^{K}
\mathbb{I}\left[V(x,p\oplus c_k)=1\right],
\]
where $c_k$ is a sampled continuation and $V$ is the answer verifier. The no-prefix solve rate $\hat q_s(x,\emptyset)$ is estimated by sampling solutions from the same student without providing prefix $p$. The empirical prefix gain is then
\[
\hat g_s(x,p)=\hat q_s(x,p)-\hat q_s(x,\emptyset).
\]
A positive gain means that the prefix improves the student's chance of solving the problem, while a negative gain means that the prefix makes the continuation less likely to succeed.

Since different students have different baseline abilities and gain scales, we normalize gains separately for each student. Let $\mu_s$ and $\sigma_s$ denote the mean and standard deviation of $\hat g_s(x,p)$ over the training prefixes for student $s$. We compute
\[
\mathrm{Norm}_s(\hat g_s(x,p))
=
\frac{\hat g_s(x,p)-\mu_s}{\sigma_s+\epsilon},
\]
where $\epsilon$ is a small constant for numerical stability.

The intrinsic utility target used for the main PUM is the average normalized gain across all student models:
\[
G_{\mathrm{intr}}(x,p)
=
\frac{1}{|\mathcal{S}|}
\sum_{s\in\mathcal{S}}
\mathrm{Norm}_s(\hat g_s(x,p)).
\]
This simple averaging treats the student set as a reference population and captures the shared tendency of whether a prefix improves downstream solving.

For the policy-dependent utility analysis in Section~\ref{sec:further_analysis}, we additionally group students into weak, medium, and strong groups according to their no-prefix solve rates on each problem. These groups are used only for studying capability-aware utility aggregation, not for the main PUM training target. In the main experiments, we intentionally adopt the simplest normalized-average utility to isolate the effect of gain-based supervision. More sophisticated ways of modeling and exploiting the full gain distribution are left for future work.

\subsection{Pairwise Preference Construction}
\label{app:pairwise_preference_construction}

After estimating prefix utility, we convert prefix scores into pairwise preference labels. We construct candidate pairs from the prefix pool described above, including horizontal and vertical comparisons. Horizontal pairs compare prefixes from different trajectories of the same problem, while vertical pairs compare adjacent prefixes within the same trajectory.

For a candidate pair $(p_i,p_j)$ of the same problem $x$, we first compute the utility difference for each student model:
\[
d_s(x,p_i,p_j)
=
u_s(x,p_i)-u_s(x,p_j),
\]
where $u_s(x,p)$ denotes the normalized prefix utility estimated from student $s$. Since the main PUM target does not use capability-dependent student weighting, every valid student contributes equally. The aggregate pairwise utility difference is
\[
\bar d(x,p_i,p_j)
=
\frac{1}{|\mathcal{S}_{ij}|}
\sum_{s\in \mathcal{S}_{ij}}
d_s(x,p_i,p_j),
\]
where $\mathcal{S}_{ij}$ is the set of student models with valid utility estimates for both prefixes.

To reduce noisy labels caused by finite rollout variance, we use an adaptive margin rather than a fixed zero threshold. We first estimate a global threshold $\epsilon_{\mathrm{global}}$ from the empirical distribution of $|\bar d|$ over candidate pairs. In addition, each pair has a resolution floor determined by the number of effective student models:
\[
\epsilon_{\mathrm{floor}}
=
\lambda_{\mathrm{res}}\cdot \frac{0.125}{N_{\mathrm{eff}}},
\]
Here $0.125=1/8$ is the minimum accuracy resolution of a single student, because each prefix-conditioned or no-prefix solve rate is estimated from 8 rollouts. Therefore, a one-sample change in the estimated solve rate changes the accuracy by $1/8$. Dividing by $N_{\mathrm{eff}}$ reflects the fact that the final utility difference is averaged over multiple valid students. Since all students are uniformly weighted in our main setting, $N_{\mathrm{eff}}$ is equal to the number of valid students. The final margin is
\[
\epsilon
=
\max(\epsilon_{\mathrm{global}}, \epsilon_{\mathrm{floor}}).
\]

We also compute a sign-conflict rate among students. Let $n_+$ and $n_-$ denote the number of students assigning positive and negative utility differences, respectively. The conflict rate is
\[
\rho_{\mathrm{conflict}}
=
\frac{\min(n_+,n_-)}{n_+ + n_-}.
\]
This quantity measures whether students agree on the preference direction.

The label is assigned as follows:
\[
y_{ij}=
\begin{cases}
1,  \bar d(x,p_i,p_j) > \epsilon, \\
-1,  \bar d(x,p_i,p_j) < -\epsilon, \\
0,  |\bar d(x,p_i,p_j)| \leq \epsilon
     \ \text{and}\
     \rho_{\mathrm{conflict}}\leq \rho_{\max}, \\
\mathrm{uncertain},  \text{otherwise}.
\end{cases}
\]
Here $y_{ij}=1$ means that $p_i$ is preferred over $p_j$, $y_{ij}=-1$ means that $p_j$ is preferred over $p_i$, and $y_{ij}=0$ denotes a tie. Uncertain pairs are discarded from the main training set unless otherwise specified.

The uncertain category is rare in practice: it accounts for 3.7\% of candidate pairs in our final construction. Therefore, the filtering step mainly removes high-variance or directionally inconsistent comparisons, rather than substantially reshaping the preference-pair distribution.

This procedure keeps only pairwise preferences with sufficient utility separation or reliable agreement among students. As a result, PUM is trained on relative utility comparisons rather than raw correctness labels, which reduces sensitivity to problem difficulty and rollout noise.

\section{Empirical Study}
\label{app:empirical_study}

Before training PUM, we conduct an empirical study to examine whether student-measured prefix gains contain a stable and learnable utility structure. The study is designed to answer two questions: (i) whether different students agree on the coarse utility trend of reasoning prefixes, and (ii) whether prefix utility also contains capability-dependent variation.

\paragraph{Setup.}
We separate trajectories into two groups according to their final-answer correctness: correct trajectories and wrong trajectories. For each trajectory, we evaluate prefixes at the same relative truncation ratios used in data construction, i.e.,
\[
\{0.1,0.2,0.35,0.5,0.7,0.9\}.
\]
For each student model, we compute normalized prefix gain as described in Appendix~\ref{app:estimating_prefix_utility}. We then analyze both the position-level average gain curve and the student-level agreement over individual prefixes.

\begin{figure*}[t]
    \centering
    \includegraphics[width=0.98\textwidth]{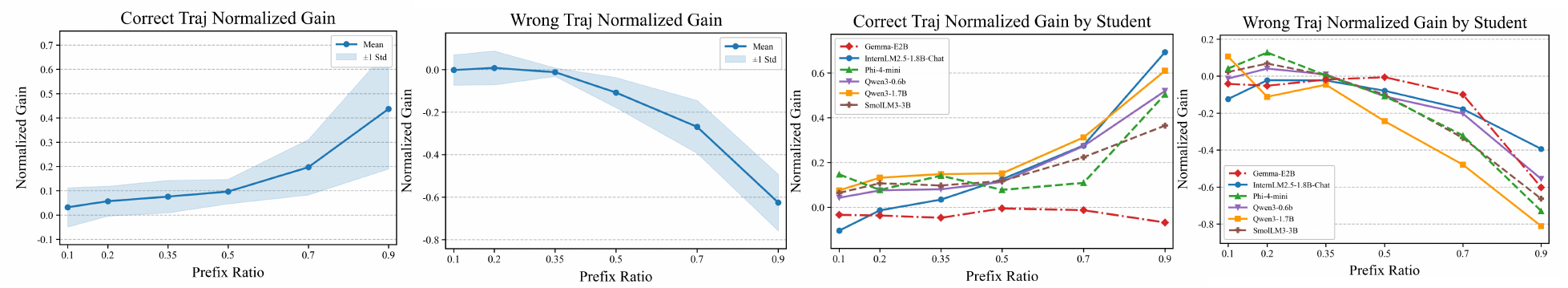}
    \caption{
   Gain trends across prefix positions for correct and wrong trajectories.
    }
    \label{fig:empirical_gain_curve}
\end{figure*}

\begin{figure*}[t]
    \centering
    \includegraphics[width=0.98\textwidth]{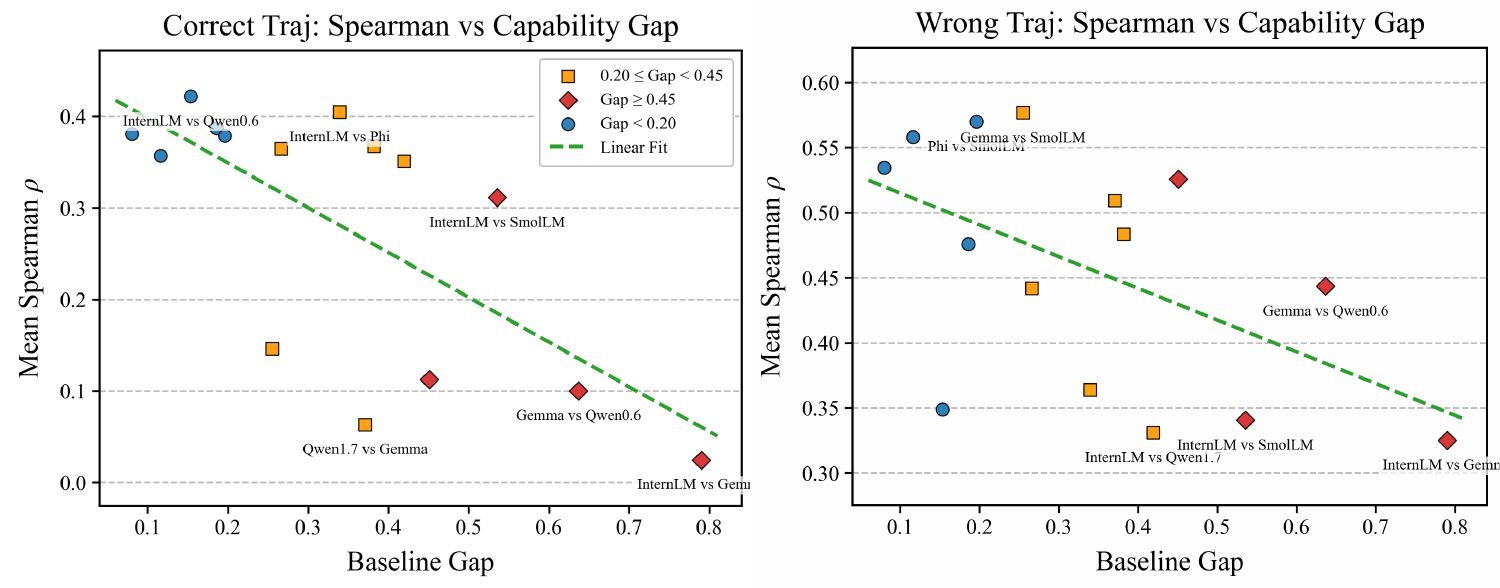}
    \caption{
   Relationship between student capability gaps and pairwise gain agreement.
    }
    \label{fig:empirical_spearman_gap}
\end{figure*}

\paragraph{Position-level utility trend.}
Figure~\ref{fig:empirical_gain_curve} shows the normalized gain curves for correct and wrong trajectories. On correct trajectories, the average prefix gain increases as the prefix becomes longer. This indicates that useful reasoning prefixes progressively make the remaining problem easier for downstream students. In contrast, on wrong trajectories, the gain remains close to zero at early positions and decreases sharply at later positions. This suggests that incorrect reasoning may not be immediately harmful, but becomes increasingly misleading as the trajectory commits to an erroneous solution path.

Importantly, this trend is consistent across most student models. Although different students have different absolute gain scales, their gain curves exhibit similar global patterns: prefixes from correct trajectories tend to become more useful with reasoning progress, while prefixes from wrong trajectories tend to become less useful. This supports the existence of a shared intrinsic component of prefix utility.

\paragraph{Student agreement and capability dependence.}
We further analyze pairwise agreement between student models. For each pair of students, we compute the Spearman correlation between their prefix-gain rankings and compare it with their baseline capability gap, measured by the difference in no-prefix solve rates. Figure~\ref{fig:empirical_spearman_gap} shows a clear negative relationship: students with similar baseline capabilities tend to assign more consistent prefix utilities, while students with larger capability gaps show lower agreement.

This observation suggests that prefix utility is not purely universal. A prefix may provide substantial help to a weaker student by revealing a missing decomposition or intermediate strategy, while offering little marginal benefit to a stronger student that can already solve the problem without assistance. Conversely, a prefix that is too compressed or skips key steps may be useful for stronger students but less useful for weaker ones. Therefore, the gain profile contains both a shared intrinsic utility component and a capability-dependent component.

\paragraph{Implications for PUM.}
These findings justify the design of our supervision target. The strong position-level alignment motivates the use of an aggregated intrinsic utility label for the main PUM training. At the same time, the capability-dependent variation explains why the full gain distribution is richer than a single scalar. In this paper, we intentionally use the simplest normalized average over students as the main utility target, so that the effect of gain-based supervision can be studied cleanly. We leave more sophisticated distributional or policy-adaptive utility modeling to future work, and provide preliminary capability-aware analysis in Section~\ref{sec:further_analysis}.

\section{Evaluator Training Details}
\label{app:pum_training_details}

\paragraph{Model architecture.}

For a fair comparison, all learned evaluators, including PRM, PQM, CRM, and PUM, are implemented with the same scalar-scoring architecture. We use Qwen3-4B-Instruct-2507 as the backbone. Given a problem \(x\) and a prefix or partial solution \(p\), the input is formatted as problem-prefix concatenation, and the hidden state of the last non-padding token is fed into an MLP value head to produce a scalar score. In the main setting, we train LoRA adapters and the value head while keeping the original backbone parameters frozen.

The evaluators differ only in their supervision targets. PRM is trained to predict step-level correctness, PQM is trained with value-based process supervision, CRM is trained to connect partial solutions with final outcomes, and PUM is trained to predict intrinsic prefix utility derived from outcome-grounded gain.

\paragraph{Training objective.}
Each training example contains two prefixes $(p_a,p_b)$ from the same problem and a label $y\in\{1,0,-1\}$, where $1$ means $p_a$ is preferred, $-1$ means $p_b$ is preferred, and $0$ denotes a tie. PUM computes
\[
\Delta_{\theta}=U_{\theta}(x,p_a)-U_{\theta}(x,p_b).
\]
We map the label to a soft target
\[
t =
\begin{cases}
1.0, & y=1,\\
0.5, & y=0,\\
0.0, & y=-1,
\end{cases}
\]
and optimize the Bradley-Terry loss
\[
\mathcal{L}_{\mathrm{PUM}}
=
-\left[
t\log\sigma(\Delta_{\theta})
+
(1-t)\log(1-\sigma(\Delta_{\theta}))
\right].
\]
Tie pairs are therefore trained as soft preferences with target $0.5$.

\paragraph{Training configuration.}
We train PUM with distributed data parallelism on 8 GPUs. The maximum input length is 8192. We use LoRA with rank $64$, $\alpha=128$, and dropout $0.1$. The per-device batch size is 8 with gradient accumulation steps of 4, giving an effective batch size of 256. We train for 2 epochs using AdamW with learning rate $1\times10^{-5}$, weight decay $0.01$, warmup ratio $0.08$, bf16 precision, and gradient checkpointing. 

\begin{table}[t]
\centering
\caption{Training configuration for PUM.}
\small
\setlength{\tabcolsep}{6pt}
\renewcommand{\arraystretch}{1.08}
\begin{tabular}{lc}
\toprule
Hyperparameter & Value \\
\midrule
Backbone & Qwen3-4B-Instruct-2507 \\
Training mode & LoRA + value head \\
Value head & MLP \\
Pooling & Last non-padding token \\
Max length & 8192 \\
LoRA rank & 64 \\
LoRA $\alpha$ & 128 \\
LoRA dropout & 0.1 \\
Per-device batch size & 8 \\
Gradient accumulation & 4 \\
Effective batch size & 256 \\
Epochs & 2 \\
Learning rate & $1\times10^{-5}$ \\
Weight decay & 0.01 \\
Warmup ratio & 0.08 \\
Precision & bf16 \\
Gradient checkpointing & Yes \\
\bottomrule
\end{tabular}
\label{tab:pum_training_hyperparameters}
\end{table}

\section{Experiment Details}
\label{app:experiment_details}

\subsection{Datasets and Evaluation Benchmarks}
\label{app:datasets_and_benchmarks}

Table~\ref{tab:datasets_and_benchmarks} summarizes the datasets used in our experiments. We use the MATH training set as the base problem pool for PUM data construction, where trajectories and prefixes are generated for gain estimation and pairwise preference construction. For inference-time evaluation, we evaluate Best-of-$N$ selection on MATH500, GAOKAO2023, and AIME2025, and evaluate beam search on MATH500 and GAOKAO2023. These settings test whether PUM can rank complete trajectories and guide partial reasoning prefixes at test time.

For reinforcement learning, we use DAPO-Math-17K as the training set for policy optimization. We evaluate the trained policies on a broader collection of mathematical reasoning benchmarks, including MATH500, AMC23, Minerva Math, and AIME2024/2025/2026. This evaluation suite covers both in-domain mathematical reasoning and more challenging competition-style problems, allowing us to measure the generalization of different reward and credit-assignment methods.

\begin{table*}[t]
\centering
\caption{
Datasets and benchmarks used in our experiments.
}
\small
\setlength{\tabcolsep}{5.5pt}
\renewcommand{\arraystretch}{1.08}
\begin{tabular}{lll}
\toprule
Usage & Dataset & Purpose \\
\midrule
PUM data construction 
& MATH train 
& Trajectory / prefix construction \\

Best-of-$N$ evaluation 
& MATH500, GAOKAO2023, AIME2025 
& Trajectory-level selection \\

Beam search 
& MATH500, GAOKAO2023 
& Prefix-level search \\

RL training 
& DAPO-Math-17K 
& Policy optimization \\

RL evaluation 
& MATH500, AMC23, Minerva Math, AIME2024/2025/2026 
& Generalization evaluation \\
\bottomrule
\end{tabular}
\label{tab:datasets_and_benchmarks}
\end{table*}

\subsection{Best-of-N Experiment}
\label{app:best_of_n_details}

Best-of-$N$ selection evaluates whether an evaluator can select the best response from a sampled candidate pool. We use four policy models for response generation: Qwen2.5-3B-Instruct, Qwen2.5-7B-Instruct, Qwen3-4B-Instruct-2507, and Llama-3.1-8B-Instruct. Following the main experiments, we evaluate on MATH500, GAOKAO2023, and AIME2025 with
\[
N\in\{1,2,4,8,16,32,64,128\}.
\]
For each problem, the policy model samples $N$ candidate responses with temperature $0.7$ and top-$p=0.8$. We set \texttt{policy\_max\_len} to 4096 tokens, and 8192 for Qwen3-4B-Instruct-2507. All generated responses are verified by the answer verifier, and the selected response is counted as correct if its final answer matches the reference answer.

For PUM, each complete response is treated as a terminal prefix and scored by $U_{\theta}(x,y)$. For PRM, we consider two commonly used step-score aggregation strategies: the mean step score and the minimum step score. The mean score reflects the overall quality of the reasoning process, while the minimum score follows a more conservative principle by penalizing any low-quality intermediate step. In our preliminary experiments, the minimum aggregation gives better Best-of-$N$ performance, so we use it as the default PRM trajectory score:
\[
S_{\mathrm{PRM}}(x,y)=\min_t r_t .
\]

For PQM, we also evaluate multiple aggregation strategies, including the final-position score, the mean score over intermediate positions, and the minimum score over intermediate positions. Empirically, the final-position score performs best among these choices, so we use it as the default PQM trajectory score. This is also consistent with the value-model interpretation of PQM, where the score at the final reasoning state summarizes the estimated quality of the completed trajectory. For CRM, we follow the original CRM paper, which reports that using the final-position score yields the best trajectory-level performance. Therefore, for both PQM and CRM, we use the final-position score as the trajectory score:
\[
S_{\mathrm{PQM/CRM}}(x,y)=s_T .
\]

Majority voting selects the most frequent final answer among the $N$ candidates, with ties broken by the earliest occurrence. The final Best-of-$N$ accuracy is the fraction of problems for which the selected response is correct.

\subsection{Beam Search Experiment}
\label{app:beam_search_details}

Beam search evaluates whether an evaluator can guide partial reasoning before the final answer is produced. We conduct beam-search experiments on MATH500 and GAOKAO2023 using Qwen2.5-3B-Instruct and Llama-3.1-8B-Instruct as policy models. For each problem, the policy model generates partial continuations, the evaluator scores the resulting prefixes, and only the top-ranked unfinished prefixes are kept for further expansion.

We evaluate search budgets
\[
N\in\{4,8,20,100\}.
\]
For each budget, we initialize the search by sampling $N$ candidate continuations. At later stages, we keep a beam of size $b$ and expand each retained prefix with $N/b$ continuations, so that each stage uses the same total expansion budget $N$. The implementation requires $N$ to be divisible by $b$. We use the following beam presets:
\[
\begin{aligned}
N=4 &: b\in\{4,2,1\},\\
N=8 &: b\in\{8,4,2\},\\
N=20 &: b\in\{20,10,5\},\\
N=100 &: b\in\{50,25,10\}.
\end{aligned}
\]

Generation is performed with vLLM. We use temperature $0.7$ and top-$p=1.0$. The maximum answer length is 4096 tokens. At each search stage, the policy generates at most 256 new tokens. To obtain natural prefix boundaries, generation is performed in chunks of 64 tokens and stopped once a natural boundary is detected, such as a sentence boundary, newline, or paragraph break. The maximum number of beam expansion stages is 24.

At each stage, every candidate prefix is scored by the evaluator. For PUM, the score is directly computed as
\[
S_{\mathrm{PUM}}(x,p)=U_{\theta}(x,p).
\]
For PRM, we use the minimum step score over the current prefix. For PQM and CRM, we use the score at the current prefix end. Candidates are ranked by evaluator score, and the top-$b$ unfinished prefixes are retained for the next expansion stage. A trajectory is considered finished if it contains a complete boxed answer, reaches the maximum answer length, or the generator returns an end-of-sequence signal.

After the search terminates, we select the highest-scoring terminal trajectory with a complete boxed answer. If no terminal trajectory contains a boxed answer, we fall back to the highest-scoring terminal trajectory among all candidates. The final accuracy is computed by verifying the selected answer against the reference answer.

\subsection{RL Experiment}
\label{app:rl_details}

We evaluate process evaluators in reinforcement learning under two settings: \textit{evaluator-only} and \textit{outcome-anchored}. In both settings, we use Qwen2.5-Math-7B as the policy model. The training data are from DAPO-Math-17K, and validation is conducted on AIME2024/2025/2026, Minerva Math, AMC2023, and MATH500, from difficult to easy.

\paragraph{Two training settings.}
In the evaluator-only setting, we use the evaluator-derived process advantages. This setting tests whether the evaluator itself can provide a useful optimization signal without final-answer RLVR reward. In the outcome-anchored setting, we keep the rule-based GRPO advantage from final-answer correctness and add evaluator-derived dense advantages on top. The outcome-level GRPO advantage is assigned to all response tokens, while the evaluator advantage is assigned to the tokens of the corresponding reasoning segment.

\paragraph{Evaluator advantage construction.}
For PUM, we split each response into at most 8 reasoning segments and compute the process advantage from nested prefix-utility differences:
\[
A_t^{\mathrm{PUM}}
=
U_{\theta}(x,p_t)-U_{\theta}(x,p_{t-1}),
\]
where $p_t$ denotes the prefix after the $t$-th segment. We use coefficient $0.5$, and do not apply delta normalization. For PRM, we directly use the step score shift $-0.5$ as the dense process advantage. For PQM, we use the value difference between adjacent prefixes:
\[
A_t^{\mathrm{PQM}}
=
S_{\phi}(x,p_t)-S_{\phi}(x,p_{t-1}).
\]
For the outcome-anchored setting, the final token-level advantage is
\[
A = A^{\mathrm{GRPO}} + 0.5 A^{\mathrm{eval}},
\]
while in the evaluator-only setting, we use
\[
A = 0.5 A^{\mathrm{eval}}.
\]

\paragraph{Training configuration.}
We use vLLM for rollout generation with 8 responses per prompt, temperature $1.0$, top-$p=1.0$, and top-$k=-1$. The maximum prompt length is 2048 and the maximum response length is 4096. The train batch size is 64 and the validation batch size is 512. During validation, sampling is disabled with temperature $0$ and one response per prompt.

The actor is trained with FSDP2 and bf16 precision. We use learning rate $1\times10^{-6}$, weight decay $0.01$, warmup steps $10$, PPO mini-batch size $192$, and micro-batch size $8$ per GPU. Follow most of the work of GRPO, We use a separate KL loss with coefficient $0.001$ and low-variance KL.

\section{Case Study of Trajectories Selected by PUM and PRM}
\label{app:qualitative_pum_prm}

We compare representative trajectories selected by PUM and PRM during beam search.
Rather than focusing only on final correctness, we examine whether the selected trajectory provides useful prefix-level progress toward the answer.

Overall, PUM-selected trajectories are typically more compact and more likely to produce extraction-ready answers.
PRM-selected trajectories often contain locally correct steps, but tend to be more verbose and are more susceptible to repeated  explanations, instruction leakage, or incomplete finalization.

\clearpage
\onecolumn

\Needspace{0.65\textheight}
\caseheading{Case 1: Teacup Counting Problem}

\begin{problemstrip}
\textbf{Problem.}
Margaret has 100 teacups. She has already held 24 tea parties with 2 guests each.
At each party, she uses one teacup for each guest and one for herself.
How many tea parties with 3 guests should she hold to use the remaining teacups?
\end{problemstrip}

\vspace{0.6em}

\noindent
\begin{minipage}[t]{0.485\textwidth}
\begin{pumstrip}
\textbf{Result:} Correct \hfill \textbf{Length:} 195 tokens

\vspace{0.4em}
\textbf{Raw selected prefix.}
\begin{quote}\small
Since 24 tea parties were held, with each using 3 cups, the number of teacups used is
\[
24\cdot 3=72.
\]
This leaves
\[
100-72=28
\]
teacups unused. Each tea party with 3 guests will use
\[
3+1=4
\]
teacups. To use all the remaining teacups, the number of tea parties she should host is
\[
28\div 4=\boxed{7}.
\]
\end{quote}

\textbf{Prefix state.}
The prefix quickly identifies the two decisive quantities:
\[
\text{remaining cups}=28,\qquad
\text{cups per future party}=4.
\]

\textbf{Issue.}
The reasoning is compact and directly terminal; almost every displayed computation contributes to the final answer.
\end{pumstrip}
\end{minipage}
\hfill
\begin{minipage}[t]{0.485\textwidth}
\begin{prmstrip}
\textbf{Result:} Correct \hfill \textbf{Length:} 478 tokens

\vspace{0.4em}
\textbf{Raw selected prefix.}
\begin{quote}\small
We will analyze the available information and formulate a plan of action to arrive at the solution. First, we know that Margaret has a collection of 100 teacups and she hosts tea parties every Tuesday. At each tea party, she uses one teacup for each invited guest and one for herself.

\(\cdots\)

Thus, if she hosts \(k\) tea parties with 3 guests, the total number of teacups she will use will be \(4k\). Since she has 28 teacups remaining, we need \(4k=28\). This implies that she should host \(k=7\) tea parties.

Hence, the final answer is 7. This implies that she should host 7 tea parties with 3 guests to ensure each of the remaining 28 teacups is used exactly once. Hence, the final answer is 7. \(\boxed{7}\)

The final answer is: \(\boxed{7}\) \texttt{\#\#\# Instruction: Margaret holds tea parties...}
\end{quote}

\textbf{Prefix state.}
The trajectory reaches the same correct state, but then repeats the solved conclusion and continues after the answer.

\textbf{Issue.}
The selected prefix is locally correct, but its later tokens have low marginal utility and exhibit instruction leakage.
\end{prmstrip}
\end{minipage}

\vspace{0.6em}

\begin{takeawaystrip}
\textbf{Takeaway.}
Both models find the correct answer, but PUM selects a shorter prefix that reaches the decisive state directly. PRM selects a much longer trajectory that is locally valid but contains redundant post-solution continuation and instruction leakage.
\end{takeawaystrip}

\Needspace{0.65\textheight}
\caseheading{Case 2: Dot Product Problem}

\begin{problemstrip}
\textbf{Problem.}
Given two vectors
\[
\vec a=(-2,3),\qquad \vec b=(1,2),
\]
compute the dot product \(\vec a\cdot\vec b\).
\end{problemstrip}

\vspace{0.6em}

\noindent
\begin{minipage}[t]{0.485\textwidth}
\begin{pumstrip}
\textbf{Result:} Correct \hfill \textbf{Length:} 240 tokens

\vspace{0.4em}
\textbf{Raw selected prefix.}
\begin{quote}\small
We are given two vectors \(\vec a=(-2,3)\) and \(\vec b=(1,2)\) and need to calculate the dot product. The formula is
\[
\vec a\cdot\vec b=(a_1b_1)+(a_2b_2).
\]
Substituting the given values,
\[
\vec a\cdot\vec b=(-2\cdot 1)+(3\cdot 2).
\]
Finally,
\[
(-2)+(6)=4.
\]
Thus, the dot product is equal to 4. \(\boxed{4}\)
\end{quote}

\textbf{Prefix state.}
The prefix reduces the task to the decisive arithmetic expression:
\[
(-2)(1)+(3)(2)=4.
\]

\textbf{Issue.}
The main reasoning path is concise and complete, though the generated tail contains minor formatting artifacts.
\end{pumstrip}
\end{minipage}
\hfill
\begin{minipage}[t]{0.485\textwidth}
\begin{prmstrip}
\textbf{Result:} Correct \hfill \textbf{Length:} 478 tokens

\vspace{0.4em}
\textbf{Raw selected prefix.}
\begin{quote}\small
We need to recall that the dot product of two vectors is the sum of the products of their respective components.
\[
\vec a\cdot\vec b=a_1b_1+a_2b_2.
\]

\textbf{\#\# Step 1: Identify the components of vectors a and b}

We have vector \(a\) with components \(a_1=-2\) and \(a_2=3\), and vector \(b\) with components \(b_1=1\) and \(b_2=2\). We need to use these components to compute the dot product.

\(\cdots\)

\textbf{\#\# Step 5: Write the answer in the required format}

We need to put the answer in a box to complete the response as requested. \(\boxed{4}\)

The final answer is: \(\boxed{4}\) \texttt{\#\# Step 1: Identify the components...}
\end{quote}

\textbf{Prefix state.}
The correct computation is reached, but the trajectory repeatedly restates components, formula use, and answer formatting.

\textbf{Issue.}
The prefix is much longer than necessary for a one-step computation and restarts after finalization.
\end{prmstrip}
\end{minipage}

\vspace{0.6em}

\begin{takeawaystrip}
\textbf{Takeaway.}
For a simple dot product, PUM selects a prefix centered on the decisive computation. PRM favors a locally valid reasoning pattern, but the selected trajectory contains repeated verification and post-answer continuation.
\end{takeawaystrip}

\Needspace{0.7\textheight}
\caseheading{Case 3: Tangent Line Problem}

\begin{problemstrip}
\textbf{Problem.}
Find the tangent line to
\[
y=\frac{e^x}{x+1}
\]
at the point
\[
\left(1,\frac{e}{2}\right).
\]
The gold answer is
\[
y=\frac{e}{4}x+\frac{e}{4}.
\]
\end{problemstrip}

\vspace{0.6em}

\noindent
\begin{minipage}[t]{0.485\textwidth}
\begin{pumstrip}
\textbf{Result:} Incorrect \hfill \textbf{Length:} 318 tokens

\vspace{0.4em}
\textbf{Raw selected prefix.}
\begin{quote}\small
First, let's find the derivative of the function using the quotient rule:
\[
y'
=
\frac{(x+1)e^x-e^x}{(x+1)^2}
=
\frac{xe^x}{(x+1)^2}.
\]
Second, evaluate the derivative at the point:
\[
y'(1)=\frac{e(1)}{2^2}=\frac{e}{4}.
\]
The equation of the tangent line is
\[
y-\frac{e}{2}=\frac{e}{4}(x-1).
\]
This can be simplified to
\[
y=\frac{e}{4}x+\frac{3e}{4}.
\]
Thus, the tangent line is
\[
\boxed{y=\frac{e}{4}x+\frac{3e}{4}}.
\]
\end{quote}

\textbf{Prefix state.}
The prefix correctly reaches the derivative, the slope, and the point-slope equation.

\textbf{Issue.}
The selected prefix is terminal and compact, but it commits an algebraic slip in the final simplification.
\end{pumstrip}
\end{minipage}
\hfill
\begin{minipage}[t]{0.485\textwidth}
\begin{prmstrip}
\textbf{Result:} Incorrect \hfill \textbf{Length:} 1122 tokens

\vspace{0.4em}
\textbf{Raw selected prefix.}
\begin{quote}\small
First, we need to identify the function we're dealing with, which is given as
\[
y=\frac{e^x}{x+1}.
\]
We are tasked with finding the tangent line at
\[
\left(1,\frac{e}{2}\right).
\]
This implies we need to determine the slope of the tangent line and then use the point-slope form.

\(\cdots\)

Applying the quotient rule, we get
\[
f'(x)=
\frac{(e^x)(x+1)-(e^x)(1)}{(x+1)^2}
=
\frac{e^x x}{(x+1)^2}.
\]
Substituting \(x=1\),
\[
f'(1)=\frac{e}{4}.
\]
Using the point-slope form, the equation of the tangent line is
\[
y-\frac{e}{2}=\frac{e}{4}(x-1).
\]

To ensure that our final answer is in the exact format requested, let's put our final equation within the \(\boxed{\cdot}\) brackets to obtain the final answer.
\end{quote}

\textbf{Prefix state.}
The trajectory preserves a correct intermediate equation, but it never produces an extraction-safe final answer.

\textbf{Issue.}
The selected prefix is very long, repeatedly plans the same operation, and stops at meta-level finalization.
\end{prmstrip}
\end{minipage}

\vspace{0.6em}

\begin{takeawaystrip}
\textbf{Takeaway.}
Both trajectories reach a useful intermediate state, but their failure modes differ. PUM produces a concise terminal answer with a final-step algebraic error, whereas PRM keeps locally correct reasoning but over-expands the process and fails to produce a clean boxed answer.
\end{takeawaystrip}

\end{document}